\documentclass[sigconf]{acmart}
\renewcommand\footnotetextcopyrightpermission[1]{} % removes footnote with conference information in first column
\usepackage{tabularx}  % 提供tabularx环境
\usepackage{booktabs}  % 提供\toprule等命令
\usepackage{array}     % 提供>{}等列格式修饰符
\AtBeginDocument{%
  }

\begin{document}

\title{Omni-Dish: Photorealistic and Faithful Image Generation and Editing for Arbitrary Chinese Dishes}

\author{
% \authoroneline{ % 单行作者模式
Huijie Liu\textsuperscript{1,2}, 
Bingcan Wang\textsuperscript{1}, 
Jie Hu\textsuperscript{1}, 
Xiaoming Wei\textsuperscript{1}, 
Guoliang Kang\textsuperscript{2}
}
%}
\affiliation{
%\textsuperscript{1} Meituan \\
%\textsuperscript{2} Beihang University \\
\textsuperscript{1} Meituan \country{China}\\
\textsuperscript{2} Beihang University
\country{China}
}

\begin{abstract}
Dish images play a crucial role in the digital era, with the demand for culturally distinctive dish images continuously increasing due to the digitization of the food industry and e-commerce.
In general cases, existing text-to-image generation models excel in producing high-quality images; however, they struggle to capture diverse characteristics and faithful details of specific domains, particularly Chinese dishes.
To address this limitation, we propose \textbf{Omni-Dish}, the first text-to-image generation model specifically tailored for Chinese dishes. 
We develop a comprehensive dish curation pipeline, building the largest dish dataset to date. 
Additionally, we introduce a recaption strategy and employ a coarse-to-fine training scheme to help the model better learn fine-grained culinary nuances.
During inference, we enhance the user’s textual input using a pre-constructed high-quality caption library and a large language model, enabling more photorealistic and faithful image generation.
Furthermore, to extend our model’s capability for dish editing tasks, we propose Concept-Enhanced P2P.
Based on this approach, we build a dish editing dataset and train a specialized editing model.
Extensive experiments demonstrate the superiority of our methods.
\end{abstract}

\begin{CCSXML}
<ccs2012>
   <concept>
       <concept_id>10010147.10010178.10010224</concept_id>
       <concept_desc>Computing methodologies~Computer vision</concept_desc>
       <concept_significance>500</concept_significance>
       </concept>
   <concept>
       <concept_id>10002951.10003227.10003251.10003256</concept_id>
       <concept_desc>Information systems~Multimedia content creation</concept_desc>
       <concept_significance>500</concept_significance>
       </concept>
   <concept>
       <concept_id>10010147.10010178.10010224.10010240.10010243</concept_id>
       <concept_desc>Computing methodologies~Appearance and texture representations</concept_desc>
       <concept_significance>300</concept_significance>
       </concept>
   <concept>
       <concept_id>10010147.10010178.10010224.10010240.10010241</concept_id>
       <concept_desc>Computing methodologies~Image representations</concept_desc>
       <concept_significance>100</concept_significance>
       </concept>
 </ccs2012>
\end{CCSXML}

\ccsdesc[500]{Computing methodologies~Computer vision}
\ccsdesc[500]{Information systems~Multimedia content creation}
\ccsdesc[300]{Computing methodologies~Appearance and texture representations}
\ccsdesc[100]{Computing methodologies~Image representations}
%\ccsdesc[100]{Computing methodologies~Texturing}
\keywords{Generative Models, Diffusion Models, Image Generation, Image Editing, Chinese Dishes}

\begin{teaserfigure}
    \vspace{-1mm}
  \includegraphics[width=\textwidth]{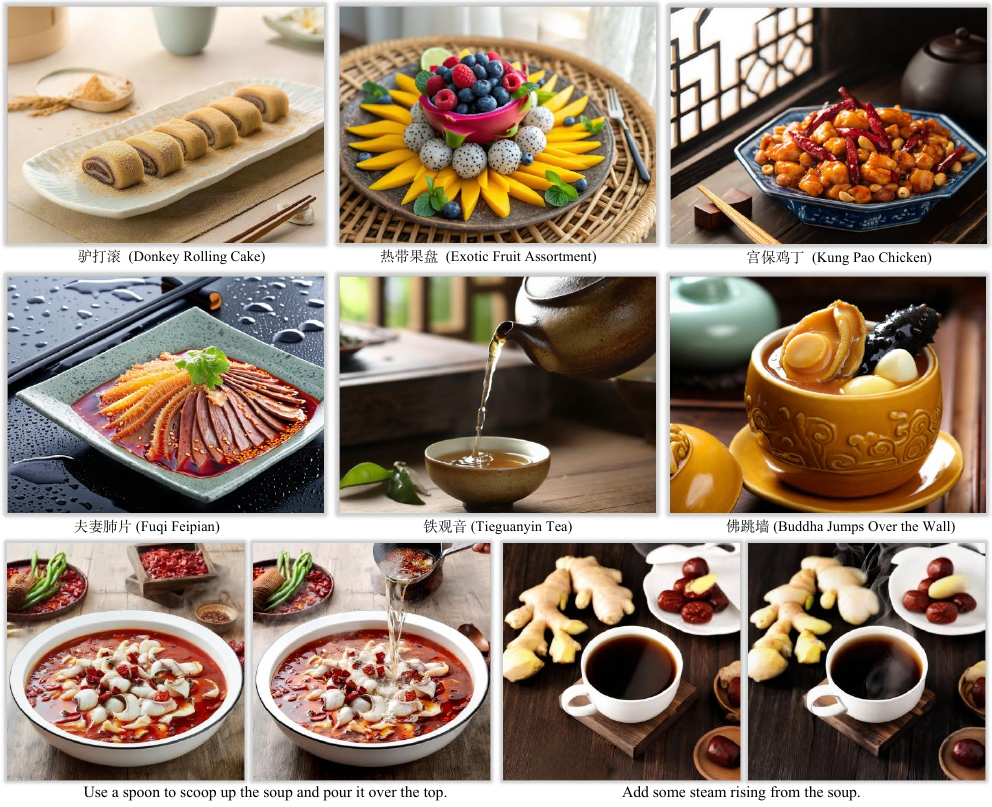}
  \vspace{-3mm}
  \caption{
  Samples demonstrate the superiority of Omni-Dish in generating and editing Chinese culinary dishes. 
  Rows 1-2 show the faithful generation results from Omni-Dish. Row 3 shows the dish editing capabilities.
  The project homepage is available at \textit{\textcolor{red}{https://liuhuijie6410.github.io/OmniDish/}}.
  \Description{Attractive pictures. Showcasing some images of Chinese dishes. Photorealistic sampling demonstrates the superiority of our method in generating and editing Chinese culinary dishes. 
  Detailed descriptions and authentic photos of all Chinese dishes referenced in this paper can be found in the Appendix.
  }
  }
  \label{fig:teaser}
\end{teaserfigure}

\maketitle
\begin{figure}[tbp]
  \centering
  \vspace{-0.5mm}
  \includegraphics[width=\linewidth]{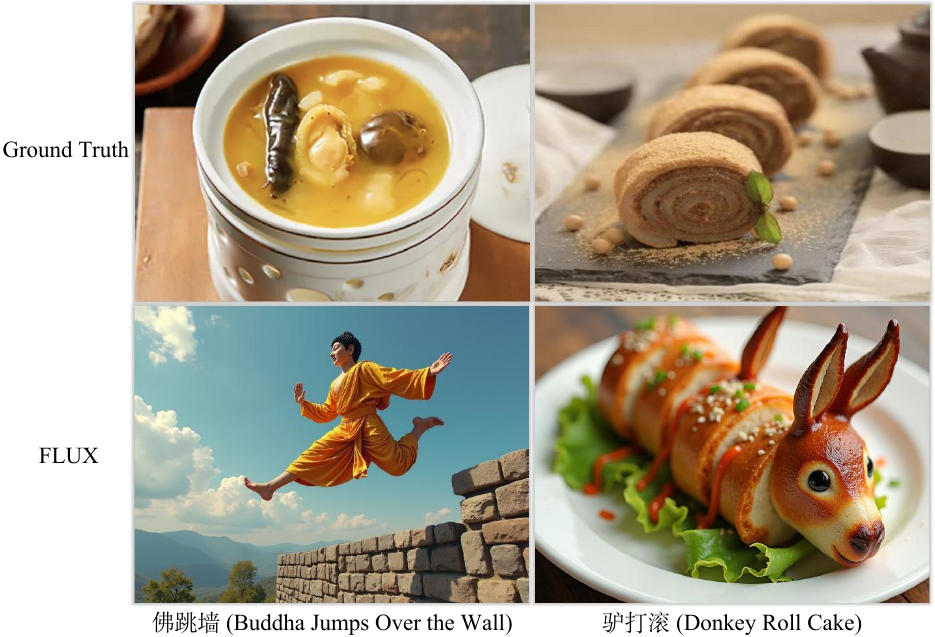}
  \vspace{-5mm}
  \caption{Existing methods face challenges in generating photorealistic and faithful images of arbitrary Chinese dishes. Row 1 shows reference images that are real photographs.
  \Description{Existing methods face challenges in generating photorealistic and faithful images of arbitrary Chinese dishes. Row 1 shows reference images that are real photographs.
  The image illustrates that the existing methods do not understand that "Buddha Jumps Over the Wall" is a Chinese dish and cannot generate details such as the texture of "Donkey Rolling."
  }
  }
  
  \vspace{-4mm}
  \label{fig:flux}
\end{figure}
\begin{figure*}[t]
 \centering
 \includegraphics[width=\linewidth]{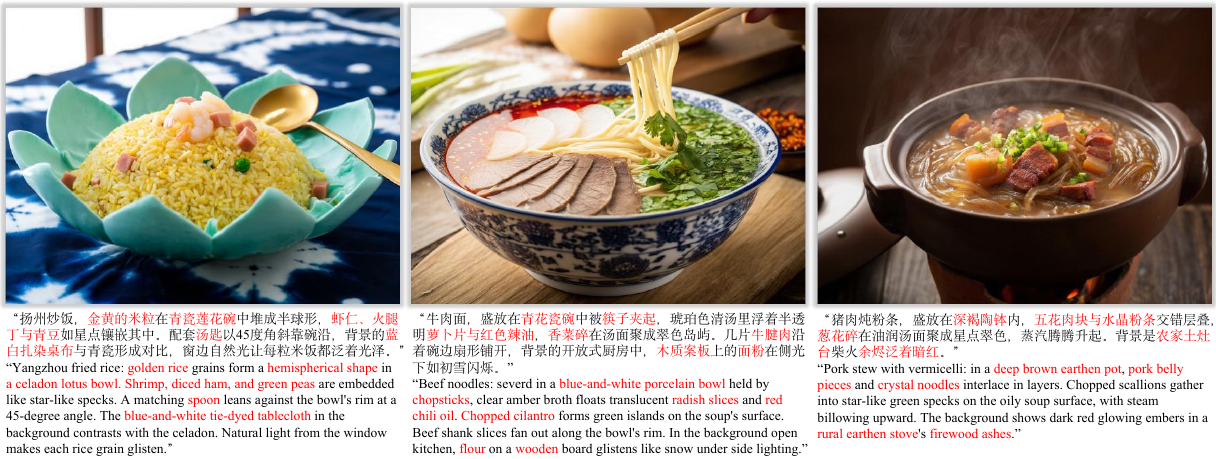} 
 \vspace{-5mm}
 \caption{\Description{Nuanced descriptions help Omni-Dish generate faithful dish images, while also endowing it with fine-grained instruction-following capabilities.}
 Nuanced descriptions not only help Omni-Dish generate faithful dish images, but also endowing it with fine-grained instruction-following capabilities.
}
\vspace{-3mm}
 \label{fig:instruct}
\end{figure*}

\vspace{-2mm}
\section{Introduction}
Dishes are closely connected to daily human experiences, and dish images are becoming an important medium in the digital era. 
High-quality dish images are particularly indispensable in daily lives. 
Consequently, generating photorealistic and faithful dish images through text input has a significant impact on multiple industries.

As shown in Fig.~\ref{fig:flux}, existing text-to-image models~\cite{rombach2022high,huang2024dialoggen,saharia2022photorealistic,podell2023sdxl,flux2024} demonstrate some limitations: (1) arbitrariness: unable to recognize dish names with Chinese cultural significance;
for example, some models can not recognize that ``Buddha Jumps Over the Wall'' is a kind of Chinese dish;  
(2) photorealism and faithfulness: unable to capture nuanced details; 
for example, they are unable to clearly render the details and textures of ``Donkey Roll Cake''.

To overcome this limitation, we propose \textbf{Omni-Dish}, a text-to-dish generation model, designed for arbitrary photorealistic and faithful dish generation.
To achieve \textbf{arbitrary dish generation}, including various niche dishes, we build an unparalleled large-scale dataset. 
We collect 100 million dish name-image pairs from China's largest catering website, subsequently implementing a meticulously designed data curation pipeline.
The pipeline employs Large Language Models (LLMs)~\cite{achiam2023gpt, glm2024chatglm, qwen2, qwen2.5}, Vision Language Large Models (VLLMs)~\cite{qwen2.5-VL,Qwen-VL,Qwen2VL, chen2024internvl}, an aesthetic scoring model, and other specialized models to perform data filtering, correction, and tagging.

However, we find that scaling up the dataset is insufficient for \textbf{photorealistic and faithful generation}.
Thus, we propose a recaption and rewriting strategy.
For recaption, we find that it is not suitable to directly use VLLMs to caption dish images as in previous methods~\cite{ramesh2022hierarchical,dalle3}, because they struggle to accurately identify the subtle elements in dish images.
Therefore, we first use LLMs~\cite{cai2024internlm2} to describe the dishes.
These descriptions are then paired with images and fed into VLLMs for recaption, which acquires more faithful descriptions.
With these recaptions, we implement a coarse-to-fine training pipeline. 
In the initial phase, the model is trained with dish images and dish names (without VLLMs' recaptions) to learn foundational dish concepts.
In the subsequent phase, dish images accompanied by high-quality recaptions are leveraged to capture fine-grained representations.
During inference, we enhance user input with a pre-constructed high-quality caption library, and further rewrite the captions with LLMs to improve the quality of generated images.
Our recaption and rewriting strategy not only enables the model to generate faithful dish images, but also improves its ability to follow instructions, as shown in Fig.~\ref{fig:instruct}.

Building upon our dish generation model, we further develop a \textbf{dish editing} pipeline that demonstrates the generation model's capacity to enable diverse downstream tasks.
Existing image editing methods~\cite{hui2024hq, huang2024smartedit,brooks2023instructpix2pix,zhang2024hive,zhao2024ultraedit} struggle to achieve dish editing, because (1) editing ingredients in dishes is often very subtle, making it difficult to achieve these fine-grained changes;
(2) some editing types for dishes are highly customized, as shown in row 3 of Fig.~\ref{fig:teaser}.

Following ~\cite{brooks2023instructpix2pix}, we construct a dataset using Prompt-to-Prompt (P2P)~\cite{hertz2022prompt}. 
%However, as shown in Fig.~\ref{}, 
We find that there exists a trade-off in the attention replacement steps of P2P, as shown in Fig.\ref{fig:p2p}. 
Extensive step replacement leads to compromised consistency in generated data pairs, while limited step replacement results in compromised consistency. 
Thus, we propose Concept-Enhanced P2P, which leverages the generation model to generate concept-specific images (not image pairs) and fine-tunes the generation model using these generated images to amplify its generation capability of the specific concept.
Then, we use P2P to generate image pairs with the fine-tuned model.
In addition, we construct data pairs for the addition and removal of ingredients in dishes by inpainting.
After final human filtering, we acquire a dataset with high consistency and observable editing effects and use it to train a dish editing model.

In the end, we conducted extensive experiments to evaluate our approach. 
For the first time, we introduced the use of the Dish T2I similarity~\cite{t2isim} to evaluate the model.
Furthermore, we constructed a high-aesthetic dish dataset as test data for FID~\cite{heusel2017gans}. 
We also conducted comprehensive human evaluations across multiple dimensions.
Specifically, we evaluated fidelity, texture, composition, scene, lighting, and subject for dish generation, as well as the effectiveness, consistency, and aesthetics for dish editing.
Extensive experiments demonstrate the effectiveness of our method.

In summary, our contributions can be summarized as follows:

\begin{itemize}
    \item We are the first to propose an image generation model specifically for dishes, \textbf{Omni-Dish}. We introduce a novel dish data curation pipeline and a recaption and rewriting strategy for training and inference, capable of generating arbitrary photorealistic and faithful dish images.
    \item We extend Omni-Dish's capability for supporting dish editing.
    Building upon Omni-Dish, we present Concept-Enhanced P2P for constructing the first open source dish editing dataset, which enables the training of editing models. 
    \item  We conducted extensive experiments, including automated metrics and carefully designed human evaluations, which demonstrate the superiority of our approach.
\end{itemize}
\section{Related Work}
\textbf{Image Generation.}
Text-to-Image generation aims to generate images conditioned on text input. Previous researchers focused on (GANs)~\cite{goodfellow2014generative, karras2019style, Karras2019stylegan2, arjovsky2017wasserstein}, which have gradually been replaced by more advanced models~\cite{ramesh2022hierarchical, saharia2022photorealistic,dhariwal2021diffusion}.
The Denoising Diffusion Probabilistic Model~\cite{ho2020denoising} proposed diffusion models based on U-Net~\cite{ronneberger2015u}, which inspired some methods such as Stable Diffusion~\cite{rombach2022high, podell2023sdxl}. 
Recently, some models such as PixArt~\cite{chen2023pixart, chen2024pixart} and FLUX~\cite{flux2024} replace U-Net with DiT~\cite{peebles2023scalable}.
However, these models, trained with English prompts, struggle to process Chinese dish names.
Even Chinese-capable models~\cite{kolors, huang2024dialoggen} such as Seedream 2~\cite{gong2025seedream} and Cogview~\cite{cogview4, zheng2024cogview3} have difficulty generating faithful Chinese dish images.

\textbf{Instruction-Based Image Editing.}
Instruction-based image editing involves altering images based on instructions, with the key challenge being the construction of high-quality datasets for model training.
HQ-Edit~\cite{hui2024hq} constructs datasets by having DALL·E 3~\cite{dalle3} create diptychs. 
However, these diptychs often exhibit poor consistency.
AnyEdit~\cite{yu2024anyedit} uses SAM, inpainting models~\cite{li2022mat} and PIH~\cite{wang2023pih} to construct a dataset.
Some other methods~\cite{huang2024smartedit, brooks2023instructpix2pix,zhao2024ultraedit} use Prompt-to-Prompt (P2P)~\cite{hertz2022prompt} to generate image pairs.
P2P represents a prevalent paradigm in image editing, although it inherently lacks image manipulation capabilities for existing photographs.  
Most editing methods~\cite{sheynin2024emu,richardson2024pops,yu2024anyedit,lee2023language, zhang2023magicbrush,fu2023guiding,mao2025ace++}  incorporate diverse designs for datasets, but often lack specialized data for dish editing.
\begin{figure*}[t]
  \centering
  \includegraphics[width=\linewidth]{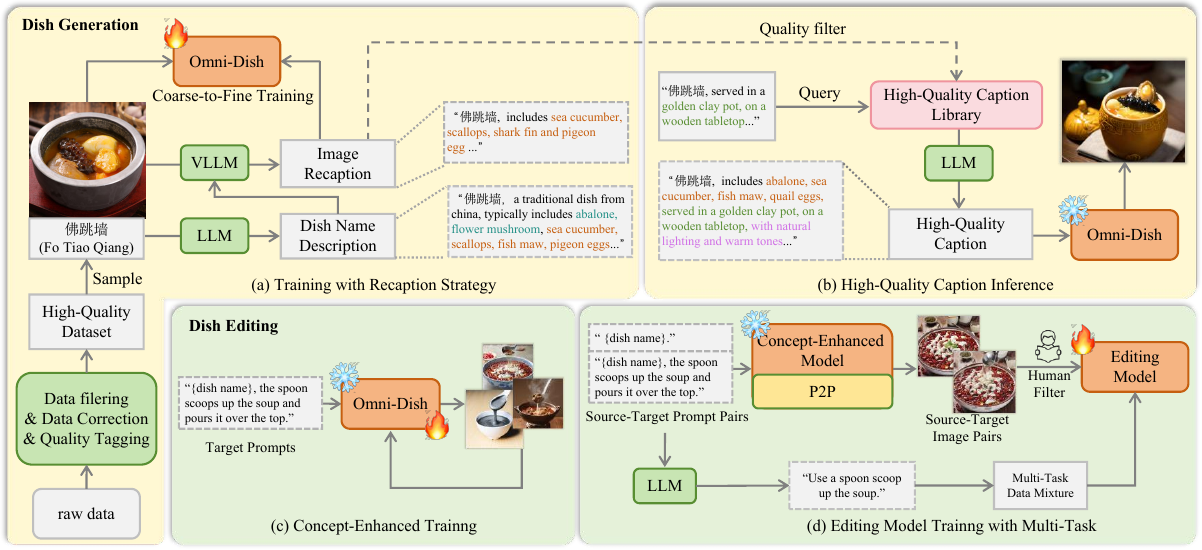}   
  \vspace{-3mm}
  \caption{Overview of our method. In the yellow block, (a) with the dish curation and recaption, the coarse-to-fine strategy is applied to train Omni-Dish; (b) high-quality captions are obtained from a pre-constructed library and rewritten by large language models for inference. In the green block, (c) the Concept-Enhanced P2P approach is introduced to build the dish editing dataset; (d) a dish editing model is trained through a multi-task data mixture.
  \Description{Method overview. On the one hand, this figure represents the dish generation approach, where we employ a coarse-to-fine training strategy with dish recaption. During inference, high-quality captions are obtained using a pre-construct caption library and LLM rewriting. 
  On the other hand, this figure shows dish editing pipeline, where we propose Concept-Enhanced P2P to construct dish-editing data and train a dedicated image-editing model for culinary dishes.}
  }
  \vspace{-3mm}
  \label{fig:overview}
\end{figure*}

\vspace{-4mm}
\section{Omni-Dish: Text-to-Dish Generation}
\label{generation model}
In this section, we propose \textbf{Omni-Dish}, the first text-to-dish generation model.
Text-to-Dish generation refers to generating photorealistic and faithful dish images based on textual dish names.
With our main emphasis on Chinese dishes, Omni-Dish can accept Chinese text input.
%It is worth noting that Omni-Dish can also generate dishes from other countries.
We present our methodology in three aspects: data curation, model training and inference.

\vspace{-1mm}
\subsection{Data Curation Pipeline}
\label{sec:data curation}
To fully cover all Chinese dish concepts, we collect 100 million data entries from China’s largest catering website to build the raw dataset, which is composed of dish name-image pairs.

\textbf{Data Filtering.} 
We first use VLLMs to detect and filter out images containing text, watermarks, and human hands.
Subsequently, we use a dish detection model~\cite{wang2022ingredient} to localize dish items through bounding box annotations, where samples with bounding boxes exceeding image boundaries (indicating incomplete dish presentation) are rigorously excluded.

\begin{figure}[tbp]
  \centering
  \includegraphics[width=\linewidth, alt={Dish name correction pipeline}]{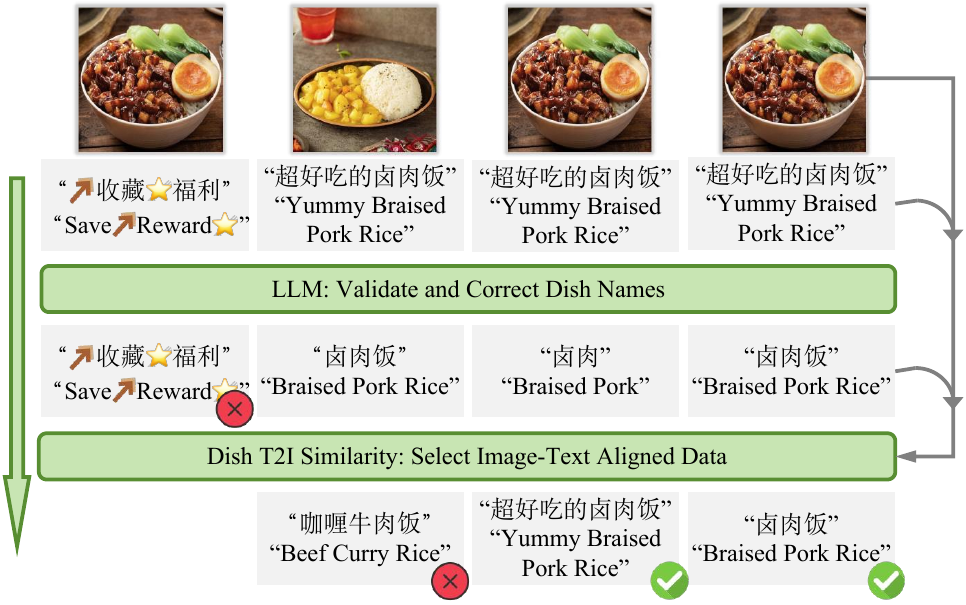}
  \vspace{-8mm}
  \caption{Dish name correction by two steps. 
  For details, refer to Data Correction in Sec.~\ref{sec:data curation}.
  \Description{Dish name correction by two steps. Step 1: Utilize LLMs to determine if the dish name describes an actual dish item. 
  Non-descriptive names are discarded. For non-standard dish names, LLMs generate a standardized correction. 
  Step 2: Calculate the DishSim.~\cite{t2isim} for both the raw name and corrected name against the dish image. If both scores fall below a predefined threshold, discard the data pair. Otherwise, retain the name (raw or corrected) with the higher similarity score.}
 }
 \vspace{-5mm}
  \label{fig:data}
\end{figure}

\textbf{Data Correction.}
The data correction pipeline is illustrated in Fig.~\ref{fig:data}.
To tackle the issue of prevalent noisy dish names, we implement LLMs to:
(a) filter out irrelevant entries that lack semantic relevance to dishes (Column 1 of Fig.~\ref{fig:data}); 
(b) correct overly descriptive but valid names (Columns 2-4 of Fig.~\ref{fig:data}).
Unfortunately, we find that the dish names corrected by LLMs are not always the correct ones we desire (Column 3, Fig.~\ref{fig:data}). 
To address this limitation, we propose to implement dish text-to-image similarity (DishSim.)~\cite{wang2022ingredient} to quantitatively evaluate the semantic alignment between candidate dish names (original vs. corrected) and their corresponding images. 
The naming variant that demonstrates the superiority of DishSim. is selected as the final correct dish name.

\textbf{Data Tagging.}
\label{data tagging}
We annotate the data with multi-dimensional tags encompassing four aspects: aesthetic quality, tableware, background, and camera angle. For example: ``served in a white ceramic bowl, placed on a brown wooden tabletop, high aesthetic quality, 30-degree shooting angle.''
These tags are used for model training as extra textual input to enhance instruction-following capabilities.
\vspace{-2mm}
\subsection{Faithful-Enhanced Training and Inference}
\textbf{Two-Stage Recaption Strategy.}
\label{sec:recaption}
Unlike most common images encountered in daily life, dish images demand greater fidelity to capture faithful details such as ingredients and textures.
We observe that models struggle to learn dish-specific nuances and textures when relying solely on dish names as textual input. 
To overcome this difficulty, a straightforward approach is to employ vision large language models (VLLMs)~\cite{qwen2.5-VL,Qwen-VL,Qwen2VL, chen2024internvl} for image recaption.
However, VLLMs often fail to infer granular information about fine-grained material composition or cooking methods directly from visual input. 
Consequently, we first employ a Large Language Model (LLM)~\cite{achiam2023gpt, glm2024chatglm, qwen2, qwen2.5} to generate comprehensive generic descriptions of dishes using dish names (rather than dish name-image pairs). 
Then, VLLMs can generate fine-grained and high-fidelity recaptions for dish images based on the introduction provided by the LLMs as a prior.

\textbf{Coarse-to-Fine Training.}
The training objectives can be decoupled into two components: dish concept learning and image quality enhancement. 
During the concept learning phase, we find that directly training the model with dish names and captions led to convergence difficulties. 
Therefore, we first train the model using standalone dish names combined with tags generated during data curation. 
In the subsequent phase, we incorporate more complex recaptions.
In the image quality enhancement phase, we employ manually annotated ultra-high-quality data for fine-tuning to improve the aesthetic quality. 
Furthermore, we use annotated human preference data for Direct Preference Optimization (DPO)~\cite{rafailov2023direct} to enhance the stability of image generation.
In summary, we employ a coarse-to-fine training strategy as illustrated in Tab.~\ref{multi-stage}.
\begin{table}[tbp]
  \centering
  \small
  \caption{
  Coarse-to-fine training. ``Tags'' refers to textual description across 4 dimensions (Sec.~\ref{data tagging}). 
  All data in Stage 4 and Stage 5 undergo manual annotation.
}
\vspace{-2mm}
\begin{tabularx}{0.48\textwidth}{@{} 
    l @{\hspace{1.2em}} 
    >{\arraybackslash}p{5.75cm}  
    >{\centering\arraybackslash}p{0.9cm}  
    @{}}
    \toprule
    Stage & Sample & Resolution\\
    \midrule
    Stage 1 & Dish name + Tags & 512 \\
    Stage 2 & Dish name + Tags + Recaption & 512\\    
    Stage 3  & Dish name + Tags + Recaption & 1024 \\
    Stage 4 & Dish name + Tags + Recaption (Ultra-High Quality) & 1024    \\
    Stage 5  & Human Preference Data (for DPO) & 1024\\
    \bottomrule
  \end{tabularx}
\label{multi-stage}
\end{table}

\textbf{High-Quality Caption Inference.} 
When users interact with the model, they often cannot provide rich fine-grained textual input like recaptions mentioned before, but instead only input a dish name with minimal description. 
This does not fully utilize the model's capability to generate details learned from recaptions.
Therefore, we filter high-quality image data from the training set and generate recaptions using the aforementioned method to pre-construct a high-quality caption library, which contains data entries composed of dish names, high-quality captions, and CLIP embeddings.
The library covers nearly all Chinese dishes with multiple captions for each dish.
During inference, given a user-input dish name, we first query the library for captions associated with that dish and then calculate the CLIP similarity.
After that, we feed both the caption with the highest similarity and the user's input text to the LLMs, instructing them to rewrite the caption to align with the user's description.
In this way, we obtain high-quality dish captions that meet user requirements, as illustrated in Fig.~\ref{fig:instruct}. 
In addition, we supplement these captions with high quality tags (e.g., ``high aesthetic quality'', ``high definition''; see Sec.~\ref{data tagging}).
Ultimately, these captions will be fed into Omni-Dish as inference inputs.

\textbf{Model Structure.}
Omni-Dish adopts the FLUX~\cite{flux2024} architecture with 7B parameters, while keeping the VAE of FLUX.1-dev~\cite{flux2024} frozen and optimizing the denoising backbone. 
To support the Chinese language, we use Qwen2.5-7B~\cite{qwen2.5} as a text encoder. 
Although our implementation primarily focuses on this architecture, the proposed method is broadly applicable to various diffusion model structures such as U-Net~\cite{ronneberger2015u} and DiT~\cite{peebles2023scalable}.

%%%%%%%%%%%%%%%%%%%%%%%%%%%%%%%%%%%%%%%%%%%%%%%%%%%%
\begin{figure}[b]
  \centering
  \vspace{-2mm}
  \includegraphics[width=\linewidth]{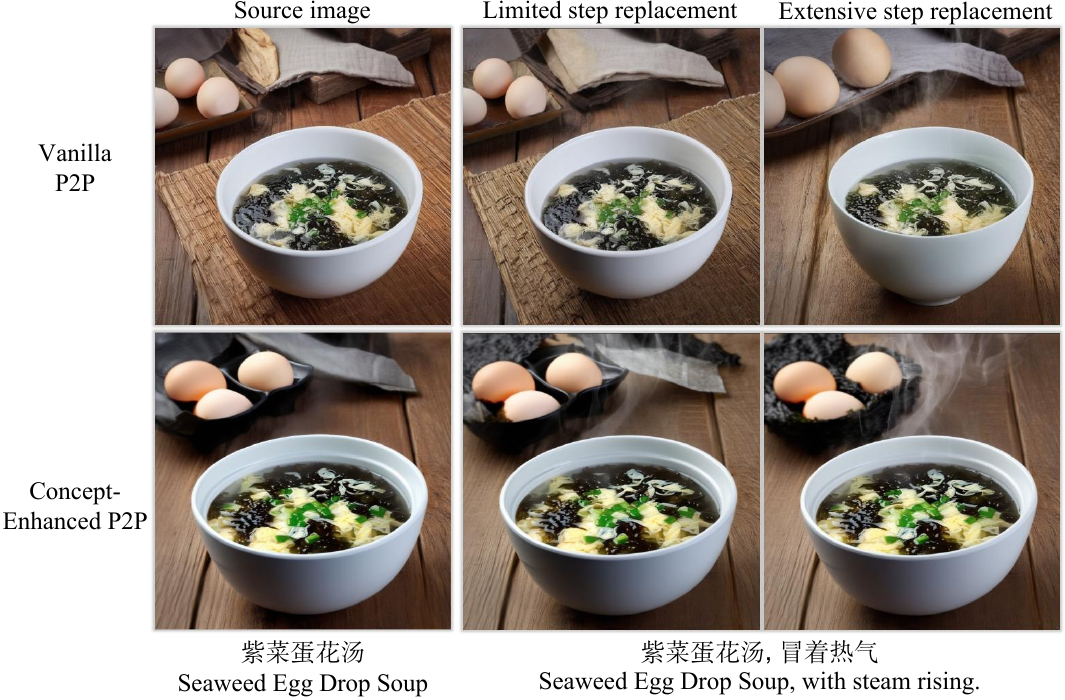}    
  \vspace{-2mm}
    \caption{Concept-Enhanced P2P can enhance editing effects while maintaining consistency.
    \Description{This picture shows Concept-Enhanced P2P can enhance editing effects while maintaining consistency.
    Without using Concept-Enhanced P2P, the generated images either have almost no steam or exhibit poor consistency between the source and edited images.
    After applying Concept-Enhanced P2P, the edited image maintain high consistency with the source image while also producing clear and realistic steam.
    }
    }
    \label{fig:p2p}
    \vspace{-2mm}
  %\caption{Introduction to Chinese cuisine, the left-side image is from the internet.}
\end{figure}

%%%%%%%%%%%%%%%%%%%%%%%%%%%%%%%%%%%%%%%%%%%%%

\begin{figure*}
  \centering
  \includegraphics[width=\linewidth]{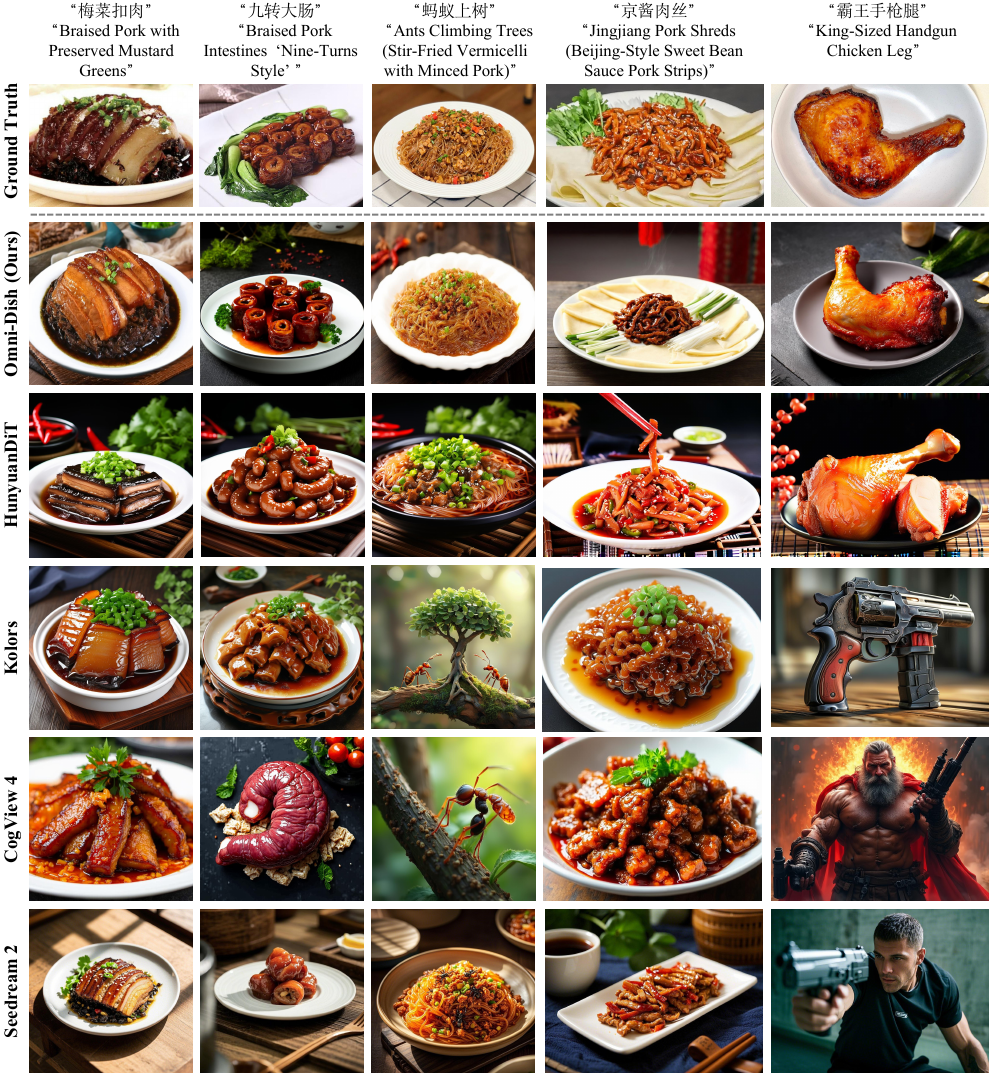}    
  \caption{Visual comparison of different models in dish generation. Row 1 displays authentic photos for readers unfamiliar with the dish, with detailed Chinese dish descriptions available in the appendix. Images in the left three columns have a resolution of 1024×1024 pixels, whereas images in the right two columns are 1024×768 pixels. 
  \Description{Visual comparison of different models in text-to-dish generation. This image presents a comparison between the results generated by the author's method and those produced by other baseline methods.
  The first row displays authentic photos for readers unfamiliar with the dish, with detailed Chinese dish descriptions available in the appendix. The images in the left three columns have a resolution of 1024×1024 pixels, while those in the right two columns are 1024×768 pixels. }
}
  \label{fig:gen}
\end{figure*}

\section{Instruction-Based Dish Editing}
\label{edit model}
Instruction-based dish image editing is more challenging compared to general editing tasks. 
Existing methods~\cite{brooks2023instructpix2pix, zhang2024hive, hui2024hq, zhang2023magicbrush, huang2024smartedit, mao2025ace++, yu2024anyedit} struggle with dish image editing. This is because the elements in dish images are more fine-grained, and instructions for dishes are more complex.
To further demonstrate the value of Omni-Dish as a dish foundation model, we utilize it to construct a dish editing dataset, and subsequently train an editing model tailored for dishes.

\vspace{-2mm}
\subsection{Concept-Enhanced P2P}
Prompt-to-Prompt (P2P)~\cite{hertz2022prompt} is a training-free method that directly maps text edits to the image generation process by replacing cross-attention maps. Given two paired prompts, it can generate paired images with consistent structure.
Existing instruction-based editing methods~\cite{brooks2023instructpix2pix,zhang2024hive} typically leverage LLMs to construct paired prompts (source and target prompts). Then they use these prompts to generate paired images by applying P2P to T2I foundation models, which heavily relies on the capabilities of foundation models. 
Omni-Dish can generate high-quality dish images, allowing us to use it for creating high-quality editing data.

However, we observe a trade-off between consistency preservation and editing effectiveness in P2P. 
P2P achieves its editing capability by replacing attention maps at specific timesteps during the denoising process. 
In this operation, excessive replacement of timesteps degrades output consistency, while insufficient timestep replacement limits editing flexibility.
As shown in Fig.~\ref{fig:p2p}, vanilla P2P is almost unable to generate steam when fewer steps are replaced; 
when more steps are replaced, weak steam is generated, but there are noticeable inconsistencies.
It is due to the model's shallow fitting of the ``steam'' concept. 
Therefore, we first use Omni-Dish to generate images conditioned on various types of target prompts.
Then use them to fine-tune Omni-Dish, thereby enhancing its ability to fit the target concept.
To help the model better learn the concept of editing types, we also incorporate some data generated from source prompts.
Ultimately, we apply P2P to this concept-enhanced Omni-Dish, which successfully builds dataset for complex editing types.

It is worth noting that we are not solely relying on P2P to build the dataset. 
We use VLLM to identify elements in the image suitable for removal, such as peppers and coriander, then use GroundedSAM~\cite{kirillov2023segany,ren2024grounded,liu2023grounding} to find the mask of the element and remove the element by inpainting.
We utilize these data bidirectionally to construct both ``remove'' and ``add'' type editing data.
To ensure the quality of the training data, all data undergo human filtering.
%which required 90 person-days of effort.
\begin{table*}[tbp]
  \caption{Comparison  of Chinese-Compatible T2I models. The introduction to the metrics can be found in Sec.~\ref{sec:metrics} and Appendix.
 We generate 3,000 samples for professional annotation teams to label over 7 days to ensure the credibility of human evaluation.
}
\vspace{-1mm}
  \begin{tabularx}{1.0\textwidth}{@{} 
    >{\raggedright\arraybackslash}m{2.4cm}
    >{\centering\arraybackslash}p{1.6cm}%1
    >{\centering\arraybackslash}p{1.6cm}%2
    >{\centering\arraybackslash}p{1.5cm}
    >{\centering\arraybackslash}p{1.5cm}
    >{\centering\arraybackslash}p{1.85cm}
    >{\centering\arraybackslash}p{1.5cm}
    >{\centering\arraybackslash}p{1.5cm}
    >{\centering\arraybackslash}p{1.5cm}@{}}
    \toprule
    \multicolumn{1}{c}{} & 
    \multicolumn{2}{c}{Automatic Evaluations} & 
    \multicolumn{6}{c}{Human Evaluations} \\
    \cmidrule(lr){2-3} \cmidrule(lr){4-9}
    \multicolumn{1}{l}{Methods} & FID$\downarrow$ & DishSim. $\uparrow$ & Fidelity $\uparrow$  & Texture $\uparrow$ & Composition $\uparrow$ & Scene $\uparrow$ & Lighting $\uparrow$& Subject $\uparrow$ \\
    \midrule
    HunyuanDiT~\cite{huang2024dialoggen} & 32.85 & 0.5909 & 2.027 & 2.354 & 2.943 & 2.524 & 2.711 &2.545\\
    Kolors~\cite{kolors}     & 31.78 & 0.5710 & 2.087 & 2.834 & 2.860 & 2.828 & 2.972 & 2.873\\    
    CogView4~\cite{cogview4}   & 27.46 & 0.5038 & 1.789 & 2.747& 2.795 & 2.863 & 2.854 & 2.807 \\
    Seedream2.0~\cite{gong2025seedream}  & 19.51 & 0.6236 & 2.265& 2.905 & 2.882 & 2.866  & \textbf{2.982} & 2.901  \\
    Omni-Dish(Ours)  & \textbf{14.96} & \textbf{0.7004} & \textbf{2.691} & \textbf{2.947} & \textbf{2.990} & \textbf{2.891}   & 2.920 & \textbf{2.913} \\
    \bottomrule
  \end{tabularx}
  \label{tab:gen}
  \vspace{-1mm}
\end{table*}
\vspace{-2mm}
\subsection{Dish Editing Training and Inference}
During training, we find that it is more critical for the model to learn the execution of editing operations (e.g., remove, add) rather than merely acquiring editing elements. 
For instance, training the editing model solely on ``add steam'' dish data (while excluding other editing-operation data) causes catastrophic failure in comprehending the general ``add'' operation, resulting in severely degraded output quality.
To mitigate this, we augment dish data with multi-task data mixture. 
Specifically, we incorporate editing data from general scenarios into the dish editing dataset to assist the model in understanding editing operations.

To incorporate the source image as an image condition, we concatenate tokens from patchified embeddings of both the noisy latent and the source image as model input. 
We augment the generation foundation model with three randomly initialized channels to process source image information.
Employing a DiT~\cite{peebles2023scalable} architecture, we train an editing model using paired data generated through the Omni-Dish framework. 
After training, the model demonstrates the capability to produce well-aligned edited images that maintain consistency with both textual instructions and the source image.
\vspace{-2mm}
\section{Experiment}
\subsection{Metrics}
\label{sec:metrics}
\textbf{Automatic Evaluations.} 
For image generation, existing benchmarks~\cite{huang2023t2i, peng2024dreambench++,hu2023tifa} often employ specified prompts to evaluate models in different dimensions, such as numeracy and relationship, which are inadequate for dish generation. 
Furthermore, since dish images typically exhibit richer visual details and we specifically focus on Chinese dish names, conventional metrics like CLIP T2I similarity~\cite{radford2021clip} and DINO T2I similarity~\cite{caron2021dino} show limited accuracy in measuring the alignment between dish names and images. 
Thus, we pioneered the use of Dish T2I Similarity, a specialized metric designed to compute the similarity between Chinese dish names and visual content, demonstrating superior performance in handling Chinese dishes.
In addition, we constructed a dedicated dataset of 22,000 high-quality dish images to measure FID-22K.

For dish editing, we used CLIP image similarity~\cite{radford2021clip} and DINO~\cite{caron2021dino} image similarity between generated images and ground truth images as evaluation metrics, following~\cite{zhang2023magicbrush, zhao2024ultraedit}.
\begin{figure}[htbp]
  \centering
  \vspace{-1mm}
  \includegraphics[width=\linewidth]{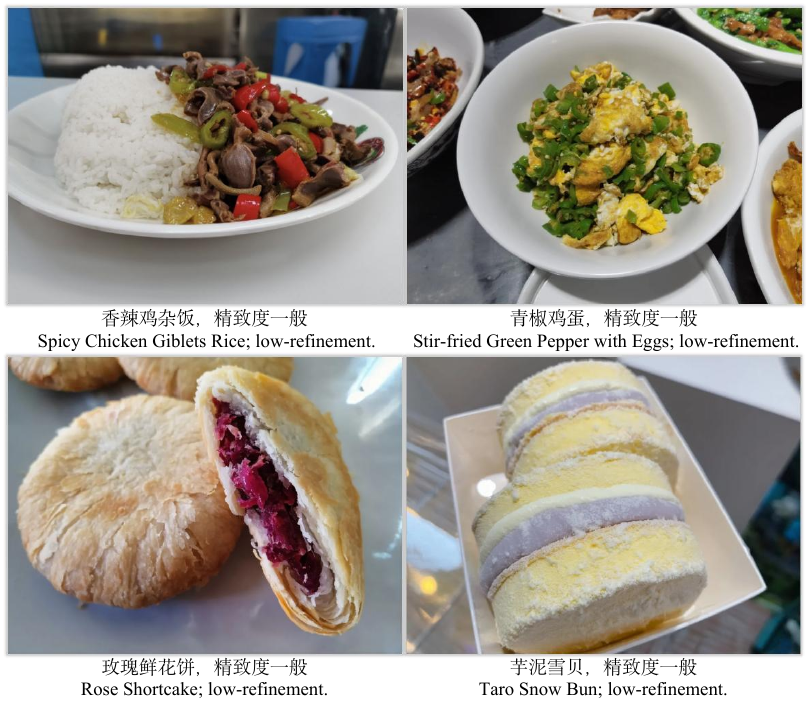}    
  \vspace{-6mm}
  \caption{With the ``low-refinement'' tag, Omni-Dish generates less polished but more realistic and authentic images.
  \Description{This image presents a set of dish images with low refinement yet remarkably authentic visual quality.}
}
\vspace{-4mm}

  \label{fig:worse}
\end{figure}
\begin{figure*}[t]
  \centering
  \vspace{-1mm}
  \includegraphics[width=\linewidth]{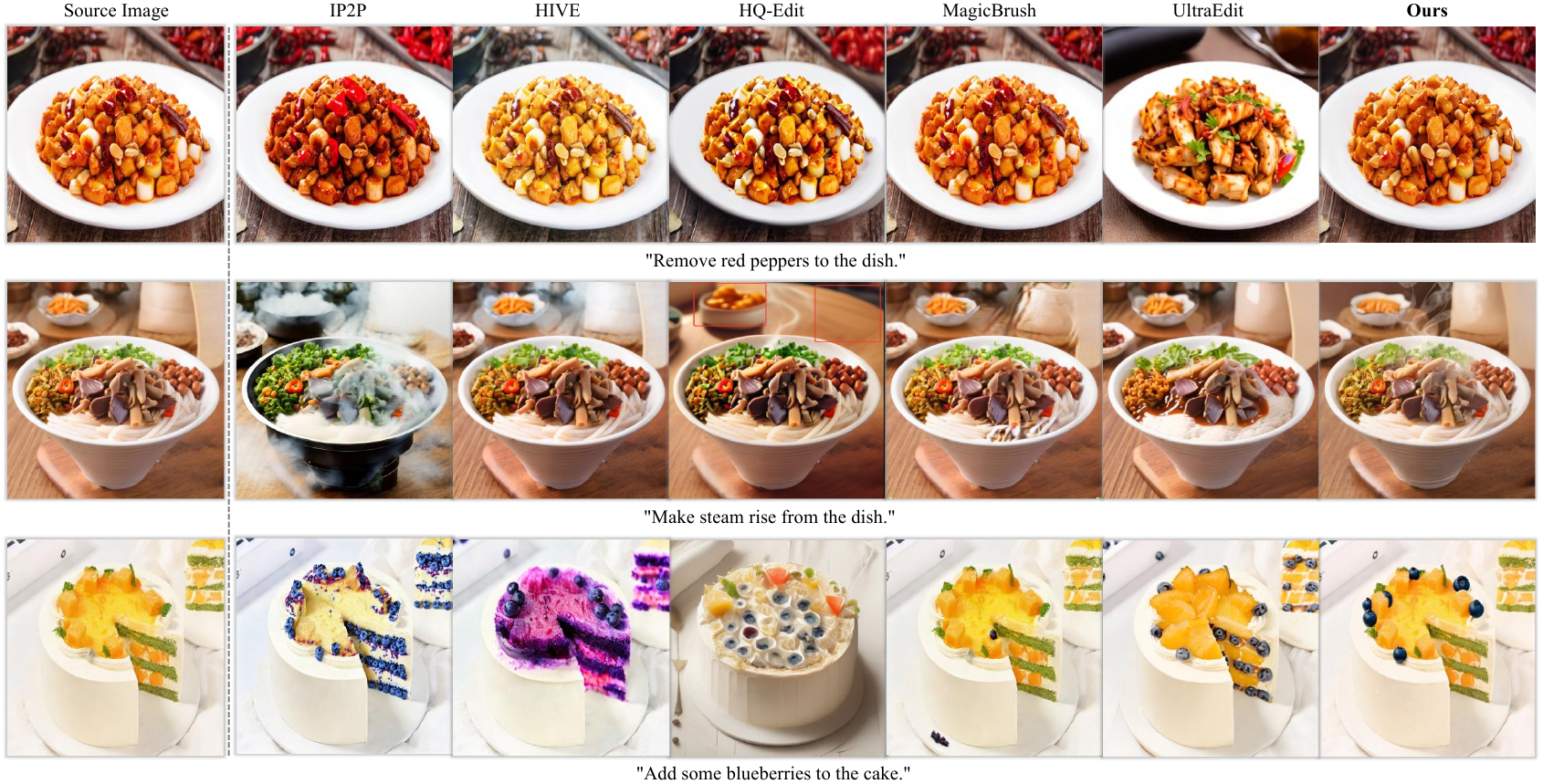}  
  \vspace{-5mm}
  \caption{Visual comparison of different models in dish editing. 
  \Description{Visual comparison of different models in dish editing. We provide three example images of dish editing, demonstrating that our method achieves superior editing results compared to other baseline approaches.}}
  \label{fig:edit}
\end{figure*}

\textbf{Human Evaluations.} 
We meticulously designed a human evaluation system for dish images. 
We briefly introduce the meaning of each evaluation dimension below, using a 3-point (1/2/3) scoring system. 
For dish generation, we evaluated models from 6 dimensions, including fidelity, texture, composition, scene, lighting, and completeness of the subject.
For dish editing, we evaluated these methods from 3 dimensions, including effectiveness (Effect.), consistency (Consist.) and aesthetics (Aes.).
We provide a detailed introduction to the measurement standards of these metrics in the Appendix.
\vspace{-2mm}
\subsection{Baselines}
For dish generation, given the lexical gap in translating culturally Chinese dish names to English, we conducted experiments with four state-of-the-art Chinese-prompt capable generative models: Kolors~\cite{kolors}, HunyuanDiT~\cite{huang2024dialoggen}, Cogview4~\cite{cogview4}, Seedream2.0~\cite{gong2025seedream}.

For dish editing, current image editing paradigms predominantly support English instructions. 
Thus, we compared our model with IP2P~\cite{brooks2023instructpix2pix}, HIVE~\cite{zhang2024hive}, HQ-Edit~\cite{hui2024hq}, and MagicBrush~\cite{zhang2023magicbrush}.
\vspace{-1mm}
\subsection{Evaluation on Text-to-Dish Generation}
The quantitative results are presented in Table~\ref{tab:gen}. Omni-Dish achieves state-of-the-art performance across most metrics, particularly in fidelity, demonstrating its capability to generate faithful arbitrary dishes. While its performance in ``Lighting'' is marginally lower than that of certain methods, it surpasses all other models..

We provide extensive visualizations in Fig.~\ref{fig:gen}.
As shown in column 3, ``Ants Climbing Trees'' is a Chinese dish composed of vermicelli and minced meat.  
Omni-Dish accurately generated both minced meat and glass noodles with authentic coloration, whereas comparative models misinterpreted the dish's nomenclature or produced unrealistic textural representations. 
Column 5 of Fig.~\ref{fig:gen} also shows similar results.
In other cases, while existing models could generate visually similar dishes, they failed to preserve faithful nuances.
For example, in column 4, ``Shredded Pork in Beijing Sauce'', typically served with cucumber strips and scallion shreds, other models omitted these essential accompaniments.

In addition, as shown in Fig.~\ref{fig:worse} with low-refinement tags, Omni-Dish can generate low-refinement but more realistic dish images.

\vspace{-1mm}
\subsection{Evaluation on Dish Editing}
In Tab.~\ref{tab:edit}, we compared the performance of several editing methods for dish images across 5 dimensions. 
Our method significantly outperforms existing models.
Although MagicBrush maintains better consistency (Consist.), its editing performance (Effect.) is poor.

We present some visual results in Fig.~\ref{fig:edit}. 
As shown in the first row (``Remove red peppers from Kung Pao Chicken.''), other methods fail to recognize what red peppers are and demonstrate nearly zero removal effect, while our method perfectly eliminates all red peppers from the dish. 
Other examples exhibit similar superior performance.

\begin{table}[tbp]
  \caption{Quantitative Comparison of Image Editing Methods.}
  \vspace{-2.5mm}
  \centering
  \small
\begin{tabularx}{0.49\textwidth}{@{} 
    >{\raggedright\arraybackslash}m{1.75cm}  % Methods column
    >{\centering\arraybackslash}p{1.0cm}  % Automatic Evaluations
    >{\centering\arraybackslash}p{1.0cm}  % Automatic Evaluations
    >{\centering\arraybackslash}p{1.0cm}  % Human Evaluations
    >{\centering\arraybackslash}p{1.0cm}  % Human Evaluations
    >{\centering\arraybackslash}p{1.1cm}  % Human Evaluations
    @{}}
    \toprule
    \multicolumn{1}{c}{} & 
    \multicolumn{2}{c}{Automatic Evaluations} & 
    \multicolumn{3}{c}{Human Evaluations} \\
    \cmidrule(lr){2-3} \cmidrule(lr){4-6}
    \multicolumn{1}{l}{Methods} & CLIP-I $\uparrow$ & DINO $\uparrow$ & Effect. $\uparrow$ & Consist.$\uparrow$ & Aes. $\uparrow$ \\
    \midrule
    IP2P~\cite{brooks2023instructpix2pix} & 0.8717 & 0.7554 & 1.61 & 1.57 & 1.43   \\
    HIVE~\cite{zhang2024hive}  & 0.8905 & 0.8096 & 1.53 & 1.65 & 1.76 \\
    HQ-Edit~\cite{hui2024hq}   & 0.7136 & 0.4541 & 1.43 & 1.17 & 1.31 \\    
    MagicBrush~\cite{zhang2023magicbrush} & 0.9437 & 0.9021 & 1.71 & \textbf{2.51} & 1.98 \\
    UltraEdit~\cite{zhao2024ultraedit} & 0.9075 & 0.8433 & 2.22 & 1.90 & 2.02 \\
    Ours & \textbf{0.9491} & \textbf{0.9057} & \textbf{2.83} & 2.30 & \textbf{2.44} \\ 
    \bottomrule
    
  \end{tabularx}
  \vspace{-3mm}
  \label{tab:edit}
\end{table}

\vspace{-1mm}
\subsection{Effect of Recaption Strategy}
\begin{figure}
  \centering
  \includegraphics[width=\linewidth]{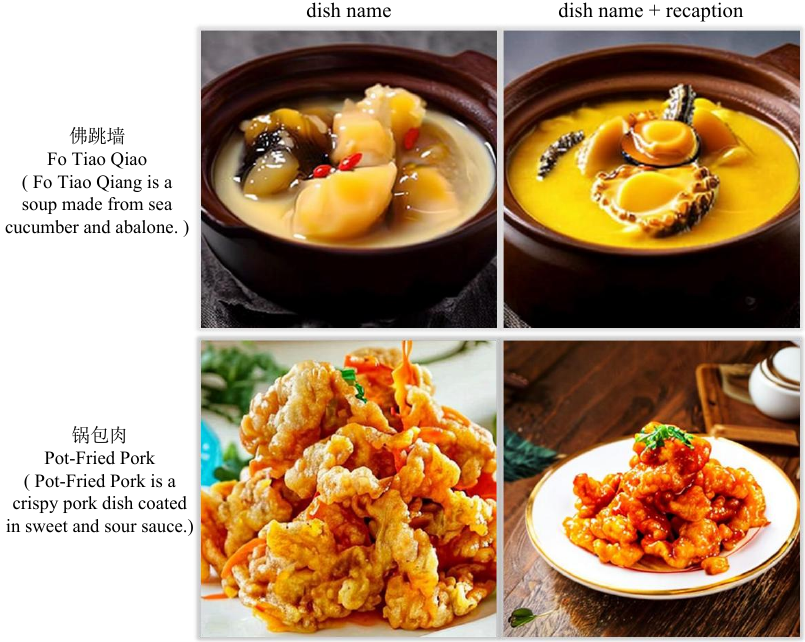}    
  \vspace{-5mm}
  \caption{Effect of our recaption strategy.
  The recaption help the model learn more details, such as the abalone in Fo Tiao Qiang and the sweet and sour sauce in Pot-fried pork. 
  \Description{Effect of our recaption strategy.
  The recaption strategy help the model learn more details, such as the abalone in Fo Tiao Qiang and the sweet and sour sauce in Pot-fried pork. 
}
\vspace{-2mm}
}
  \label{fig:ablation}
\end{figure}

We validated the impact of employing a recaption strategy on model training. 
To conserve computational resources, we conducted the experiment with limited training (1M samples). 
As illustrated in Fig.~\ref{fig:ablation}, with the recaption strategy, the contours of abalone in ``Fo Tiao Qiang'' become more distinct, while the outer layer of ``Pot-fried pork'' clearly exhibits a sweet and sour sauce coating.

\section{Conclusion}
In this paper, we propose Omni-Dish, the first image generation model specifically designed for Chinese dishes. 
The generated images exhibit not only photorealism but also faithfully capture the intricate ingredient compositions and cooking details of the dishes.
Building upon our generation model, we further develop an editing model to achieve high-quality dish editing.
Extensive experiments demonstrate the impressive effectiveness of our approach.

\clearpage
\bibliographystyle{ACM-Reference-Format}
\bibliography{sample-base}

\end{document}